\documentclass{article}

    \usepackage[preprint]{neurips_2023}

\newcommand{\myparagraph}[1]{\vspace{5pt}\noindent\textbf{#1}}
\newcommand{\modelfullnametitle}{Reconstruct-and-Generate Diffusion Model~}
\newcommand{\modelfullname}{reconstruct-and-generate diffusion model~}
\newcommand{\modelabbrname}{RnG}
\newcommand{\modelabbr}{RnG}

\usepackage{natbib}
\setcitestyle{numbers,square}

\usepackage[utf8]{inputenc} 
\usepackage[T1]{fontenc}    
\usepackage{hyperref}       
\usepackage{url}            
\usepackage{booktabs}       
\usepackage{amsfonts}       
\usepackage{nicefrac}       
\usepackage{microtype}      
\usepackage{xcolor}         

\usepackage{graphicx}
\usepackage{amstext}
\usepackage{subfigure}
\usepackage{tabularray}
\usepackage{amsmath}
\usepackage{multirow}
\usepackage{algorithm}
\usepackage{algorithmic}
\usepackage{caption}
\usepackage{wrapfig}
\usepackage{xcolor}
\usepackage[export]{adjustbox}

\title{\modelfullnametitle for Detail-Preserving Image Denoising}
\author{%
  Yujin Wang\thanks{Corresponding author} \\
  Shanghai Artificial Intelligence Laboratory\\
  \texttt{wangyujin@pjlab.org.cn} \\
 \And
  Lingen Li \\
  The Chinese University of Hong Kong \\
  \texttt{lgli@link.cuhk.edu.hk} \\
 \And
  Tianfan Xue\\
  The Chinese University of Hong Kong \\
  \texttt{tfxue@ie.cuhk.edu.hk} \\
 \And
  Jinwei Gu \\
  The Chinese University of Hong Kong \\
  \texttt{jwgu@cuhk.edu.hk} \\
}

\begin{document}

\maketitle

\begin{abstract}
Image denoising is a fundamental and challenging task in the field of computer vision. 
Most supervised denoising methods learn to reconstruct clean images from noisy inputs, which have intrinsic spectral bias and tend to produce over-smoothed and blurry images. 
Recently, researchers have explored diffusion models to generate high-frequency details in image restoration tasks, but these models do not guarantee that the generated texture aligns with real images, leading to undesirable artifacts. 
To address the trade-off between visual appeal and fidelity of high-frequency details in denoising tasks, we propose a novel approach called the \modelfullnametitle (\modelabbrname). Our method leverages a reconstructive denoising network to recover the majority of the underlying clean signal, which serves as the initial estimation for subsequent steps to maintain fidelity. Additionally, it employs a diffusion algorithm to generate residual high-frequency details, thereby enhancing visual quality. We further introduce a 
two-stage training scheme to ensure effective collaboration between the reconstructive and generative modules of \modelabbrname. To reduce undesirable texture introduced by the diffusion model, we also propose an adaptive step controller that regulates the number of inverse steps applied by the diffusion model, allowing control over the level of high-frequency details added to each patch as well as saving the inference computational cost. Through our proposed \modelabbrname, we achieve a better balance between perception and distortion.
We conducted extensive experiments on both synthetic and real denoising datasets, validating the superiority of the proposed approach.

\end{abstract}

\vspace{-12pt}
\section{Introduction}
\vspace{-8pt}

\begin{figure}[ht]
  \centering
  \includegraphics[width=0.9\textwidth]{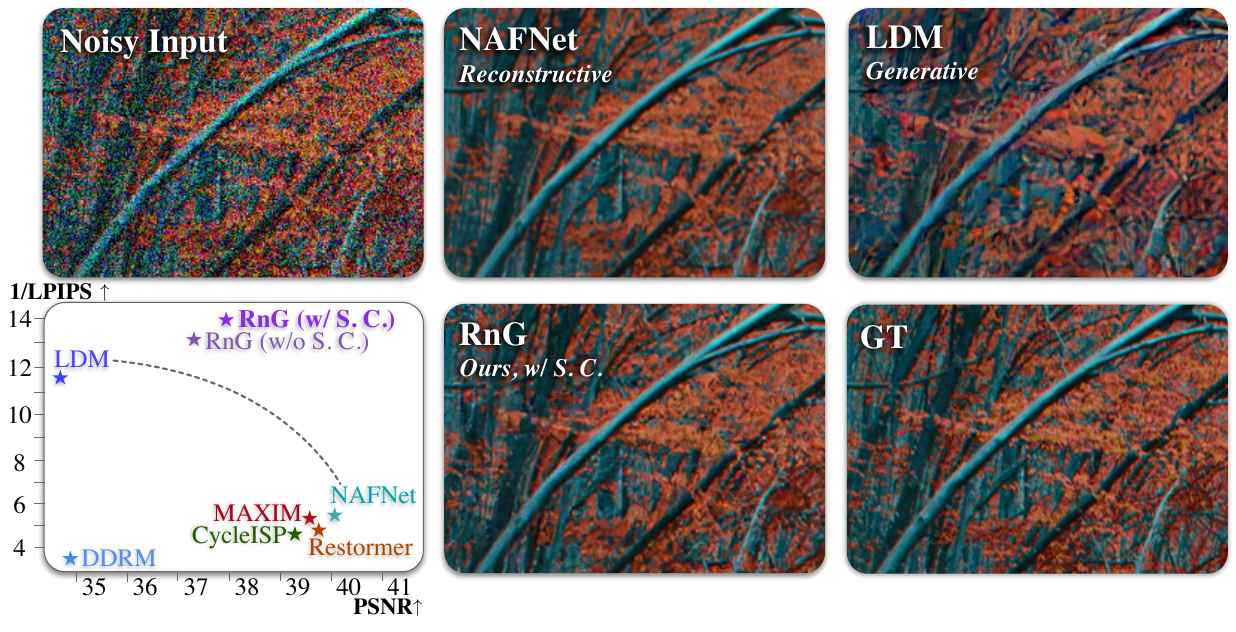}
  \caption{In this work, we proposed a Reconstructive-and-Generative (RnG) denoising model, that achieves the best trade-off between perception (measured by LPIPS) and distortion (measured by PSNR) as shown on the left. The visual result also indicates the proposed RnG produces detail-preserved denoising results. On the contrary, the reconstructive method, like NAFNet, often generates over-smoothed results (shown on the right). The generative method on the other side, sometimes outputs artificial textures that are not consistent with ground truth. The \textit{S.C.} is short for step controller of the proposed \modelabbrname.}
  \label{fig:teaser}
  \vspace{-10pt}
\end{figure}


Image denoising is a fundamental problem in computational photography, with wide applications in medical imaging, remote sensing, computer vision, and more. 
DNN-based reconstructive image denoising methods~\cite{chen2022simple, zamir2020cycleisp, tu2022maxim, zamir2022restormer} learn to reconstruct clean images from noisy observation, which are promising but intrinsically tend to produce blurry images due to their spectral bias~\cite{rahaman2019spectral, xu2019frequency} and the ill-posed nature of the denoising problem.
Recent advances in generative models~\cite{goodfellow2020generative, alsaiari2019image, divakar2017image} have demonstrated the great potential of preserving high-frequency details in the denoising process. Generative methods improve visual perception by decorating noisy areas with generated high-frequency textures. However, this improvement comes at the cost of fidelity loss. Researchers have observed that these methods occasionally introduce undesired new textures that are unrelated to the original clean images, resulting in high-frequency artifacts~\cite{lugmayr2021ntire}. In Figure~\ref{fig:teaser}, we can see an example where the generative model, Latent Diffusion Model (LDM), produces a denoised output with textures that significantly deviate from the ground truth. 
The perception-distortion trade-off curve is proposed to describe this phenomenon, and the performance of current state-of-the-art methods can be plotted as a curve that represents the current perception-distortion trade-off frontier, as shown in the bottom-left part of Figure~\ref{fig:teaser}.

Given the considerable performance of reconstructive and generative methods in distortion and perception metrics, respectively, can we further push the boundary of the current perception-distortion curve by harnessing the strengths of both approaches?

To this point, we propose a novel \modelfullname (\modelabbr) for detail-preserving image denoising, which contains a reconstructive module and a generative module, leveraging the strengths of both two types of methods. Its reconstructive module produces an initial estimation that recovers low-frequency clean images with high fidelity. This initial estimation is then enhanced by the generative module, a conditional diffusion model, to include additional residual high-frequency details. Additionally, the reconstructive module can be any DNN-based denoiser, and the NAFNet is proven to be the best candidate in our experiments. The proposed RnG diffusion model addresses the fidelity problem in common generative methods used for image-denoising tasks like DDRM and LDM, pushing the limit of the current perception-distortion curve a step forward.

Furthermore, we recognize that in some cases, such as images with simple textures like the sky, the initial estimation is already visually satisfactory, and further generated texture may actually introduce artifacts to the final result. 
As high-frequency details are incrementally injected during the iterative sampling and denoising process, it is possible for us to limit the generated high-frequency details when necessary.
Therefore, we introduce a step controller that controls the generation process by predicting the sampling step of the diffusion model for each input patch to avoid undesirable artifacts, improving perceptual performance and avoiding meaningless computational costs.
Extensive experiments demonstrate that the proposed {\modelabbrname} outperforms existing state-of-the-art models on multiple perceptual metrics with real and synthetic denoising datasets, achieving superior visual quality.

Our main contributions can be summarized as follows:
\begin{itemize}
\setlength{\itemsep}{2pt}
\vspace{-4pt}
\item[$\bullet$] We proposed the \modelfullnametitle, which alleviates the fidelity problem of generative denoising methods and conveniently achieves better perceptual quality, leading to a better trade-off between perception and distortion.
\vspace{-4pt}
\item[$\bullet$] We proposed an adaptive step controller to control the generation intensity of the {\modelabbrname} based on the observation that the increasing step in diffusion may lead to artifact, making the proposed {\modelabbrname} flexible to different inputs without any expensive retraining or fine-tuning on the diffusion model itself and further achieving higher perceptual quality and less distortion.
\item[$\bullet$] Our approach enables us to expedite the inference of diffusion-based models by utilizing initial predictions and a step controller to reduce the number of diffusion inference steps.
\end{itemize}


\vspace{-12pt}
\section{Related Work}
\vspace{-8pt}
Traditional image denoising methods are often based on image priors~\cite{aharon2006k, dong2011sparsity, elad2006image,buades2005non, dabov2007image, dabov2008image,dong2012nonlocal, gu2014weighted}. 
With the development of deep neural networks (DNNs), data-driven denoising methods achieved significant advances~\cite{jain2008natural}. DnCNN~\cite{zhang2017beyond} is one of the widely used convolutional denoising networks. Following DnCNN, many different network architectures were designed based on convolutional neural network (CNN)~\cite{chen2022simple, zamir2020cycleisp, tu2022maxim} and Transformer~\cite{liang2021swinir, zamir2022restormer}. Those methods usually adopt pixel-wise loss as a straightforward training objective and optimize the denoising model to achieve high performance in PSNR (peak signal-to-noise ratio). However, the well-known phenomenon perception-distortion trade-off~\cite{blau2018perception} indicates that PSNR and other distortion metrics only partially correspond to human perception, and high distortion metrics are usually accompanied by lower perceptual quality.
 
In addition to supervised denoising approaches, there are studies that concentrate on using unsupervised denoising methods to circumvent the need for collecting noisy-clean training pairs.
One approach to unsupervised denoising involves generating training pairs using a generative adversarial network (GAN)~\cite{chen2018image, cha2019gan2gan}. Another approach is to train a neural network solely on noisy input images~\cite{batson2019noise2self, xu2020noisy, krull2019noise2void, lehtinen2018noise2noise, quan2020self2self}.
Although no paired data is required for training, existing unsupervised denoising methods still exhibit noticeable performance gaps compared to their supervised counterparts~\cite{pang2021recorrupted}. Therefore, in this work, we focus on supervised denoising.

Recently, diffusion model~\cite{ho2020denoising, song2020score, song2019generative} has achieved state-of-the-art perceptual results in the field of image generation. There are also novel approaches~\cite{kawar2021stochastic, kawar2022denoising} for handling the image denoising task using diffusion models by samples from the posterior distribution given the noisy image. Diffusion models are employed in other restoration tasks~\cite{saharia2022image, whang2022deblurring, rombach2022high, saharia2022palette} to solve the image inverse problem. 
However, it is widely observed that diffusion models usually report lower distortion metrics~\cite{saharia2022image, whang2022deblurring}, such as PSNR, than other non-generative approaches, indicating their lack of fidelity in image restoration tasks.

\begin{figure}[t]
  \centering
  \includegraphics[width=1\textwidth]{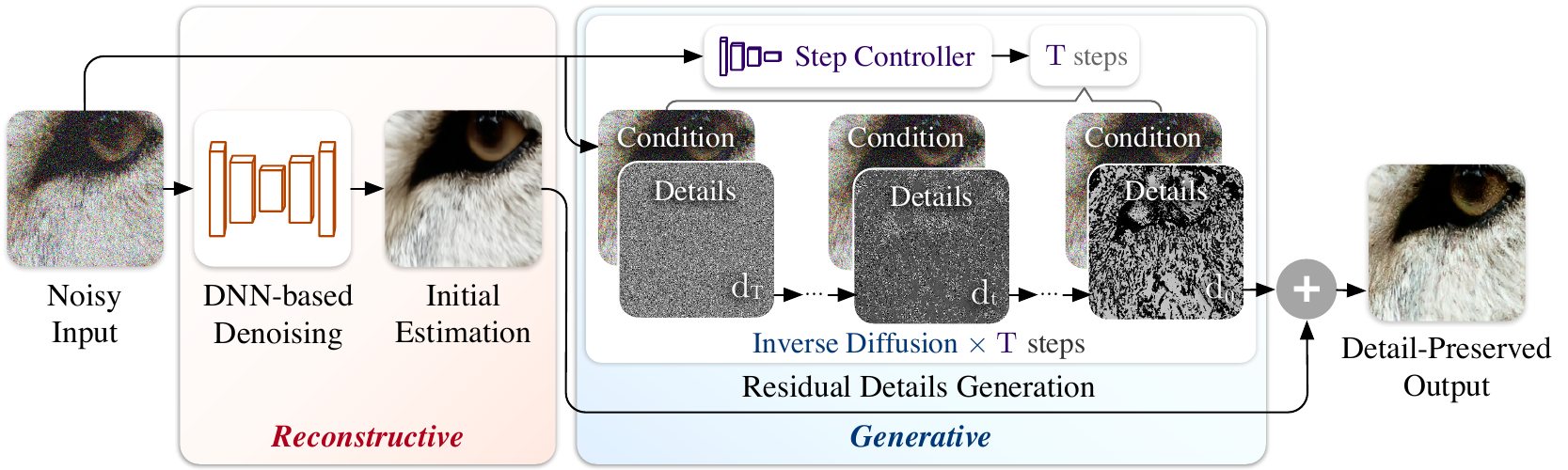}
  \caption{The proposed {\modelabbrname} consists of two key components: the reconstructive module, which utilizes a DNN-based denoiser, and the generative module, which generates residual details in inverse diffusion. The reconstructive module produces an initial estimation from the noisy input, while the generative module generates residual details conditioned on the noisy input. To further enhance perceptual performance, an additional step controller takes the noisy image as input and predicts the optimal number of inverse diffusion steps. This step controller effectively controls the generation intensity and prevents the occurrence of undesirable artifacts.}
  \label{fig:framework}
  \vspace{-10pt}
\end{figure}

\vspace{-10pt}
\section{\modelfullnametitle}
\vspace{-8pt}

In this section, we will introduce the \modelabbrname, which consists of the reconstructive module and generative module, whose framework is illustrated in Figure~\ref{fig:framework}. The {\modelabbrname} combines the merit of the reconstructive denoising method, which maintains the fidelity of the output images, and the advantage of the generative method, which preserves high-frequency details. 

Specifically, given a noisy image $x_i$, the corresponding clean image $y_i$, the reconstructive module employs a DNN-based denoiser $r_{\theta}$ to produce the initial estimation $r_{\theta}(x_i)$ from the noisy image, where $\theta$ are learnable model weights.
In addition, the generative module employs a diffusion model $g_{\theta}$ to learn the condition distribution $P(y_i - r_{\theta}(x_i) | x_i)$ and outputs the MAP (Maximum A Posteriori Estimation) result, denoted as $g_{\theta}(x_i)$. 
The final denoising output is the summation of the two parts, which is denoted as $r_{\theta}(x_i) + g_{\theta}(x_i)$. Figure~\ref{fig:framework} summarizes the overall workflow of our method.



Compared to previous diffusion models~\cite{rombach2022high, saharia2022image}, one important difference of the diffusion model used in our generative module is that our model only learns a conditional distribution of residual details given the noisy image. There are two advantages of this design. 
Firstly, the reconstructive denoising model can recover the majority of low-frequency components but may wipe out some high-frequency details, whereas the generative diffusion model can help. Secondly, current diffusion models for denoising directly learn to generate clean images from noisy observations without any initial estimation~\cite{rombach2022high, saharia2022image}, which requires the diffusion model to learn both the faithful low-frequency and the high-frequency components. On the contrary, the generative module of our proposed model focuses on learning the high-frequency details only and leaves the prediction of the low-frequency component to the reconstructive module. 
Since the reconstructive module has higher fidelity, our approach can reduce the possibility of the diffusion model producing artifacts that deviate from target clean images. 


In the following, we will first give a quick overview of the generative diffusion model and then introduce our step controller, training scheme, and optimization targets.


\vspace{-4pt}
\subsection{Generative Model}
\vspace{-4pt}
Our denoising training dataset consists of noisy-clean image pairs $\left\{{x}_{i}, {y_{i}}\right\}^{N}_{i=1} $. To train our models, we first generate noisy-detail data pairs $\left\{x_{i}, {d_{i}}\right\}^{N}_{i=1} $, where $d_i = y_i - r_\theta(x_i)$.
We then train a denoising diffusion probabilistic (DDPM) model~\cite{sohl2015deep,ho2020denoising} that estimates details $d_i$ given the denoising output of the reconstructive model $r_\theta(x_i)$.



Following the classic diffusion process setup, we define a forward Markovian diffusion process $q$ that gradually adds Gaussian noise to a clean residual detail image $d_0$ over $T$ iterations:
\begin{equation}
q(d_{t}\mid d_{t-1}){=}{\mathcal{N}({d_{t}}\mid\sqrt{\alpha_{t}}\,d_{t-1},(1-\alpha_{t})\textbf{I})},
\end{equation}
where $\alpha_t \in (0,1) \text{ for all } t = 1,...T$ and controls the variance of noise added at each step. To model the posterior distribution $p(d_{t-1} | d_0, d_t)$, we use the reparameterization trick to train a denoising network $g_\theta$ by minimizing the variational lower bound. Specifically, we take noise level ($\gamma_{t}\ {{=}}\ \prod_{i=1}^{t}\alpha_{i}$) as input to the denoising model $g_\theta$, similar to~\cite{chen2020wavegrad, song2019generative, whang2022deblurring, saharia2022image}. Please refer to the supplementary material for the detailed structure of $g_\theta$. And training the model by minimizing this objective function:
\begin{equation}
\mathrm{E}\left\vert\!\left\vert\epsilon-g_{\theta}(\sqrt{{\gamma}_{t}}d_{0}+\sqrt{1-{\gamma}_{t}}\epsilon,{\gamma}_{t}, x)\right\vert\!\right\vert^{p}_{p},
\label{eqn:expectation_for_training}
\end{equation}
where $\epsilon\sim\mathcal{N}(0,I)$ is the Gaussian noise added at each step, $x$ is a noisy image from the training dataset, and $d_0$ is the residual high-frequency detail data obtained by $d_0 = y - r_\theta(x)$.

The inference process is defined as a reverse Markovian process starting from Gaussian noise $d_T$, and iterative infers $d_t$ as:
\begin{equation}
d_{t-1}\leftarrow{\frac{1}{\sqrt{\alpha_{t}}}} (d_{t}-{\frac{1-\alpha_{t}}{\sqrt{1-\gamma_{t}}}}g_{\theta}(x,d_{t},\gamma_{t}))+{\sqrt{1-\alpha_{t}}}\epsilon_{t},
\label{eqn:inference}
\end{equation}
where $\epsilon\sim\mathcal{N}(0,I)$. Detailed derivation of Equations~\ref{eqn:expectation_for_training} and~\ref{eqn:inference} will be presented in the supplementary material.

In the inference stage, we conduct a small grid search over the noise schedule hyper-parameters and select the model with the best LPIPS score~\cite{zhang2018unreasonable}, similar to~\cite{whang2022deblurring}. We note that this inference-time hyper-parameter tuning is relatively free because it does not require retraining or finetuning the model itself.

\begin{figure}[t]
\includegraphics[width=1\textwidth]{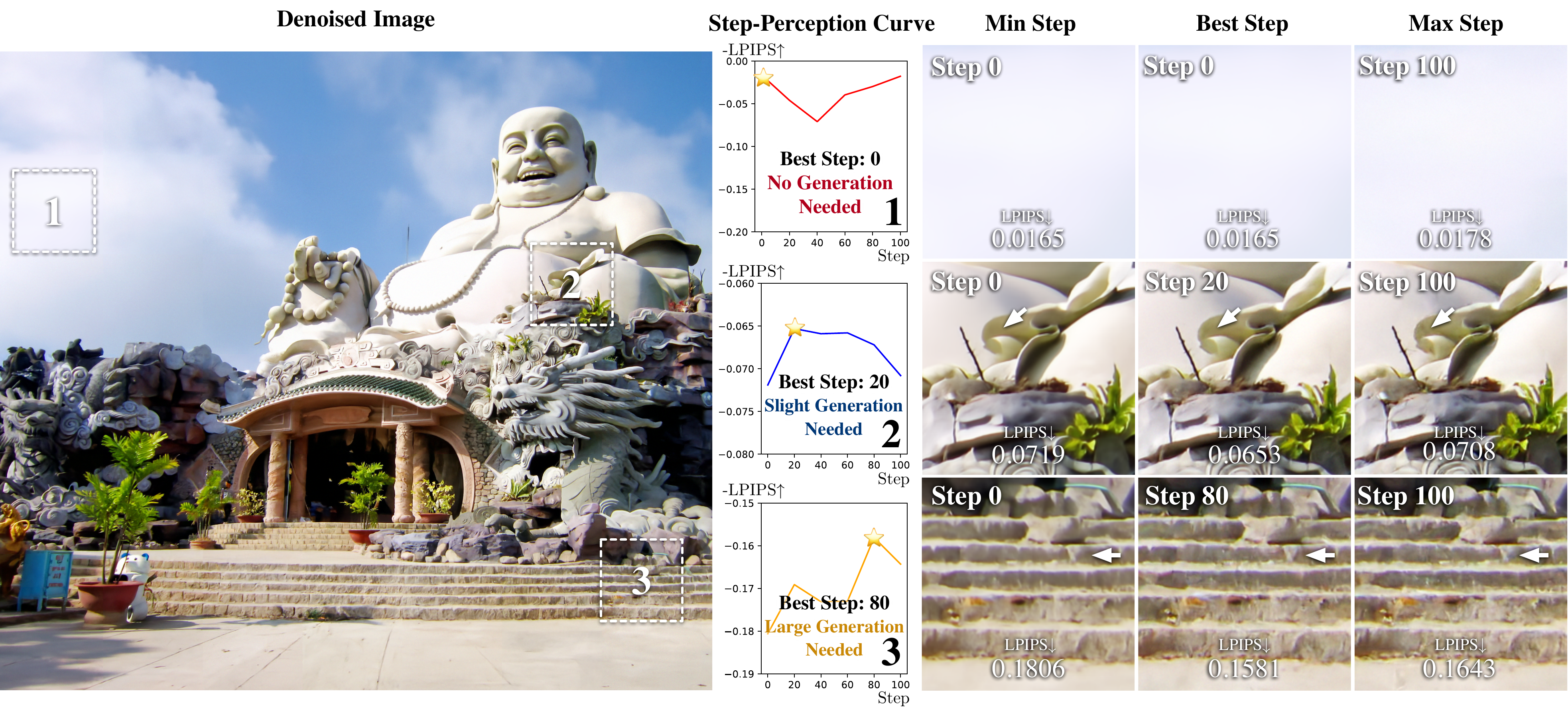}
\caption{Illustration of patches generated in best diffusion steps that are searched towards better LPIPS and three selected patches generated in min (the zero step, which is equivalent to generating nothing), best, and max steps. As we can see, different patches have their own best steps, indicating that a content-aware step controller is needed for optimal image denoising.
}
\label{fig:demo-why-do-we-need-step-controller}
\vspace{-15pt}
\end{figure}

\vspace{-8pt}
\subsection{Step Controller}
\vspace{-4pt}
To further reduce potential artifacts introduced by the generative diffusion process, we have developed a diffusion step control module that determines the optimal amount of high-frequency details to be injected into the final output. For many smooth regions like the sky, the initial estimation of the reconstructive denoising module is already close to the ground truth, and additional high-frequency details generated by the generative diffusion module may either be unnecessary or even result in artifacts, thereby compromising performance. As depicted in the first row of Figure~\ref{fig:demo-why-do-we-need-step-controller}, in a smooth sky region, the direct output of the reconstructive model ($step = 0$) is very similar to the result obtained after 100 iterations of the generative diffusion module, indicating that further generation is unnecessary. Moreover, in the second row, with a 20-step diffusion sampling, it can already recover most of the high-frequency details. Additional details added in the later diffusion sample actually result in artifacts that hurt the visual quality, and the quantitative visual measure (-LPIPS, higher is better) also drops from -0.0653 to -0.0708 when the diffusion step increases from 20 to 100. Only very few patches require a large diffusion step to recover all details, as shown in the third row. 

With this observation, it is critical to adaptively determine the diffusion for each low region. Recent work~\cite{whang2022deblurring} can conveniently traverse the Perception-Distortion curve to achieve different perceptual qualities by manually tuning the different inference steps of the diffusion model, which also proves the possibility of building a control policy of its generation intensity. However, they ignored the demand for refined strength in different texture areas and achieved different perceptual qualities only through manually tuning the inference step, which is infeasible in practice. We consider that different local regions may require different numbers of diffusion steps and design a step controller module to address this issue.

To implement this step controller, we first collect a training dataset by running a trained diffusion model on a large number of image patches, about tens of thousands of image patches. We then conduct a small grid search over the inference step ($\text{step} = [0, 10, ...,100]$) for every noisy patch and select the best inference step for each patch with minimum LPIPS score~\cite{zhang2018unreasonable} (LPIPS measures the perceptual difference between diffusion output and ground truth). The final training pairs consist of input image patches and corresponding ground truth steps for each patch obtained by this searching processing.

We then treat the step prediction as a classification problem. Specifically, we build a lightweight convolution network that is much smaller than reconstructive or generative networks. It consists of eight convolutional blocks, a fully connected layer, and a softmax activation~\cite{bridle1989training}, each convolutional block is composed of a 3 $\times$ 3 convolution layer, a batch normalization layer, and a ReLU activation. Maxpooling layers are appended after the second, the fifth, the sixth, and the eighth block respectively. Two residual connections are added:  one between the second and fourth layer, and another between the sixth and eighth layer. Finally, we optimize our network using the cross-entropy loss function. 

In the inference stage, given the original noisy image, the step controller module will predict a diffusion sampling step number for each local patch, to control the strength of the high-frequency detail generation, as shown in Figure \ref{fig:framework}. Algorithm~\ref{alg:inference} summarizes the entire inference process of the \modelabbrname~model.

In most experiments, metrics are only calculated on $256\times 256$ patches.
In the case of larger images with resolutions exceeding $256 \times 256$, we divide them into $256 \times 256$ patches with 32-pixel overlap and apply the generative component on each patch individually. When stitching the generated results together, we discard the 12 overlapped boundary pixels from each $256 \times 256$ patch while retaining the average value of the remaining 8 boundary pixels. Because our diffusion model only generates high-frequency details, no obvious boundary artifacts are not observed.
Please refer to our supplementary material for more visual results.

\vspace{-4pt}
\subsection{Training Scheme and Optimization Targets}
\vspace{-4pt}
To train the reconstructive module, we employ the following training target:
\begin{equation}
{\cal L}_{\mathrm{Reconstructive}}(\theta)= \mathrm{E}\left\vert\!\left\vert y - r_{\theta}(x)\right\vert\!\right\vert^{p}_{p}.
\end{equation}
 For the generative module, we use the following formula as the training objective:
\begin{equation}
{\cal L}_{\mathrm{Generative}}(\theta)= \mathrm{E}\left\vert\!\left\vert\epsilon-g_{\theta}(\sqrt{{\gamma}_{t}}(y - r_{\theta}(x))+\sqrt{1-{\gamma}_{t}}\epsilon,{\gamma}_{t}, x)\right\vert\!\right\vert^{p}_{p}.
\end{equation}
In the implementation, we choose $p=2$ for the reconstructive module and $p=1$ for the generative module.

\begin{algorithm}[t]
\caption{{Inference process of the \modelfullname.}}
\label{alg:inference} 
\begin{algorithmic}[1] 
\REQUIRE ~~\\ 
$x$: noisy image; $r_\theta$: reconstructive model; $g_\theta$: generate model; $s_\theta$: step controller model. 
\ENSURE ~~\\ 
$\hat{y}$: detail-preserved denoising result.
\STATE {$\hat{y}_r = r_\theta(x)$}; {$d_t \sim \mathcal{N}(0,I)$}; {$T = s_\theta(x)$}; \\
\FOR {$t \gets T $ \TO $1$}
    \STATE {$\epsilon_t \sim \mathcal{N}(0,I)$}; {$\gamma_{t}\ {{=}}\ \prod_{i=1}^{t}\alpha_{i}$}; 
    \STATE {$d_{t-1}\leftarrow{\frac{1}{\sqrt{\alpha_{t}}}} (d_{t}-{\frac{1-\alpha_{t}}{\sqrt{1-\gamma_{t}}}}g_{\theta}(x,d_{t},\gamma_{t}))+{\sqrt{1-\alpha_{t}}}\epsilon_{t}$}
\ENDFOR
\RETURN $\hat{y} = \hat{y}_r + d_0$
\end{algorithmic}
\end{algorithm}

At last, we introduce the novel two-stage training scheme for the \modelabbrname. Specifically, we first use the pixel-loss function to train the reconstructive model. When the reconstructive model converges, we freeze the parameters and then start training the generative model. Unlike most multi-module networks, we did not end-to-end fine-tune both modules together but trained them sequentially instead.
This is because as shown by previous research~\cite{xu2019frequency,rahaman2019spectral}, when training a convolutional neural network to fit a function (in this case, it is a denoising function), it often fits the low-frequency component of the function first and fits the high-frequency component in the later training iterations. If the reconstructive module and generative module are jointly optimized, they will interfere with each other and not work as expected.
Therefore, it is necessary to train our {\modelabbrname} with the two-stage scheme, letting the diffusion model focus on learning high-frequency details when training the generative module. Our experiments in Section~\ref{sec:ablation_framework} also demonstrate that it is crucial to train our {\modelabbrname} by using the two-stage training scheme.

\vspace{-8pt}
\section{Experiments}
\vspace{-6pt}

In this section, we will introduce experiments that verify the effectiveness of the proposed approach.

\myparagraph{Dataset.} We train and evaluate our models on widely used real and synthetic denoising datasets. For a fair comparison, we follow the same setup used by~\cite{chen2022simple,zamir2022restormer,liang2021swinir,zhang2018residual} and evaluate our model on:
\begin{itemize}
\setlength{\itemsep}{1pt}
\item \textbf{Real Image Denoising Dataset}. We use the Smartphone Image Denoising Dataset (SIDD)~\cite{abdelhamed2018high}, a widely used real denoising dataset. It consists of images from 10 static scenes under different lighting conditions captured by 5 representative smartphone cameras. There are 320 image pairs for training and 1280 image patch pairs for validation.
\item \textbf{Gaussian Image Denoising Dataset}. Following~\cite{liang2021swinir,zhang2018residual}, we conduct denoising experiments on several widely-used synthetic color image benchmarks (DIV2K~\cite{agustsson2017ntire} and Flickr2K ~\cite{young2014image}) with additive white Gaussian noise. Compared with testing benchmark datasets (CBSD68~\cite{martin2001database}, Kodak24~\cite{franzen1999kodak}, McMaster~\cite{zhang2011color}, Urban100~\cite{huang2015single}), DIV2K is more diverse, with 800 training images, 100 validation images, and 100 test images. The training dataset consists of images from both DIV2K and Flickr2K, and the validation dataset consists of images from DIV2K, with synthetic added Gaussian noise with a standard deviation of $\sigma \in [20,35,50]$.
\end{itemize}


\vspace{-8pt}
\myparagraph{Training Details.\label{sec:data}} 
We employ the state-of-the-art denoising method, NAFNet~\cite{chen2022simple}, as our reconstructive denoising module, adopting the same training setup similar to~\cite{chen2022simple}. Specifically, our training data consists of 128 $\times$ 128 randomly cropped patches and we adopt training-time data augmentation with random horizontal/vertical flips and 90°/180°/270° rotations, similar to ~\cite{chen2022simple}, with a batch size of 256. We use the AdamW~\cite{loshchilov2017decoupled} optimizer with a fixed learning rate of 0.0001, a weight decay rate of 0.0001, and an EMA decay rate of 0.9999. During training, following~\cite{saharia2022image, whang2022deblurring}, we use a fine-grained diffusion process with $T$ = 2000 steps and use a linear noise schedule with the two endpoints set as: 
$1 - \alpha_{0} = 1 \times 10^{-6} \text{ and } 1 - {\alpha}_{T} = 0.01$.


\myparagraph{Evaluation Metrics.} We evaluate our method on different distortion-based metrics and perceptual metrics: PSNR, SSIM~\cite{wang2004image}, LPIPS~\cite{zhang2018unreasonable}, NIQE~\cite{mittal2012making}. We use the implementation from the image quality assessment toolbox by~\cite{pyiqa}\footnote{We use the assessment toolbox provided at \url{https://github.com/chaofengc/IQA-PyTorch}.}
It should be noted that the improvement in distortion accuracy (e.g., PSNR, SSIM) is not always accompanied by an improvement in perceptual quality, which is known as the Perception-Distortion trade-off~\cite{blau2018perception}. Although an increase in perception metric will also lead to a decrease in distortion metric, one would never need to sacrifice more than $3dB$ in PSNR to obtain perfect perceptual quality. 

\begin{figure}[tb]
  \centering
  \includegraphics[width=1\textwidth]{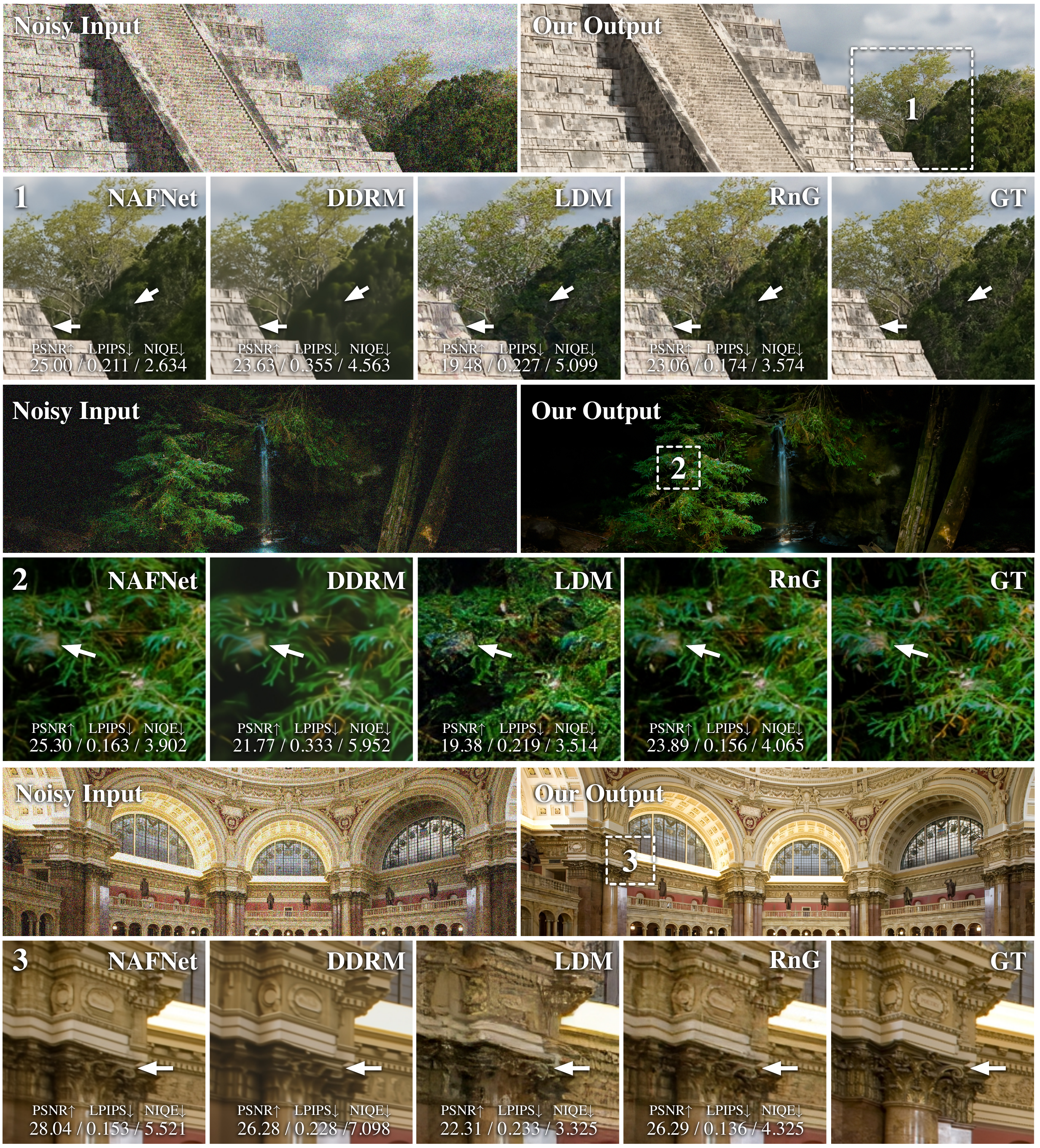}
  \caption{Visual comparison on the DIV2K dataset (zoom-in for a better view). The proposed RnG method preserves more details compared with the reconstructive method (NAFNet). Furthermore, the high-frequency details generated by RnG are closer to the ground truth, compared with pure generative methods (DDRM and LDM). The proposed RnG method also reports the best LPIPS metric. \textbf{Please refer to our supplementary material for full-resolution comparison, where differences between different methods are more obvious.}}
  \label{fig:comparison-synthetic}
\vspace{-17pt}
\end{figure}

\vspace{-4pt}
\subsection{Results on Real Noise}
\vspace{-4pt}
Table \ref{tab:sidd-result} shows denoising results on the real noise dataset, SIDD~\cite{abdelhamed2018high}. We compared our model with four reconstructive-based denoising methods (NAFNet~\cite{chen2022simple}, Restormer~\cite{zamir2022restormer}, Cycle-ISP~\cite{zamir2020cycleisp}, MAXIM~\cite{tu2022maxim}), generative-based denoising methods (DDRM~\cite{kawar2022denoising}, LDM~\cite{rombach2022high}) and GAN-based methods (NAFNet with perceptual loss~\cite{johnson2016perceptual}, PDASR~\cite{zhang2022perception}). Our model achieves the best performance across both perceptual metrics (both LPIPS and NIQE) while there are only minor drops in distortion metrics (PSNR and SSIM). Note that pure generative methods (DDRM and LDM) and GAN-Basd methods (PDASR) sacrifice more than $5dB$ in PSNR, which results in notable artifacts in the final result. Also, perceptual metrics (LPIPS and NIQE) are more consistent with our visual perception. \textbf{Please refer to our supplementary material for more visual comparison.} Furthermore, our approach outperforms diffusion-based techniques by requiring only 76 averaged inference steps to achieve the best results.



\begin{table}[t]
\small
\caption{Experimental results of ablation on training strategy, reconstructive denoising net and generative diffusion condition using SIDD dataset. The combination of the two-stage training scheme, condition with noisy image, and reconstructive module of NAFNet lead to the best settings, which is the framework of the proposed RnG. \textit{Joint Opti.} is short for joint optimization.  \textit{Inter. Supv.} stands for intermediate supervision. \textit{Init. Est.} stands for initial estimation.}
\label{framework ablation}
\centering
\resizebox{\columnwidth}{!}{
\begin{tabular}{@{}ccc|cc|cc|cc@{}}
\toprule
\multicolumn{3}{c|}{Training Scheme}         & \multicolumn{2}{c|}{Reconstructive Module} & \multicolumn{2}{c|}{Condition}  & \multicolumn{2}{c}{Metric}             \\ \midrule
{Joint Opti.}  & {Inter. Supv.} & {Two-Stage}        & {U-Net}            & {NAFNet}              & {Noisy}            & {Init. Est.}   & {PSNR↑}              & {LPIPS↓}            \\ \midrule
$\checkmark$ &              &                  & $\checkmark$     &                     & $\checkmark$     &              & 32.4438            & 0.1360            \\
             & $\checkmark$ &                  & $\checkmark$     &                     & $\checkmark$     &              & 37.2392            & 0.1185            \\
             &              & $\checkmark$     & $\checkmark$     &                     & $\checkmark$     &              & 36.7467            & 0.1091            \\
             &              & $\checkmark$     &                  & $\checkmark$        &                  & $\checkmark$ & 38.0268            & 0.1116            \\
             &              & $\checkmark$  &                  & $\checkmark$           & $\checkmark$  &              & $\textbf{38.1088}$ & $\textbf{0.0997}$ \\ \bottomrule
\end{tabular}
}
\vspace{-18pt}
\end{table}

\vspace{-8pt}
\subsection{Results on Synthetic Gaussian Noise}
\vspace{-4pt}
Table \ref{tab:div2k-result} shows denoising results on synthetic Gaussian noise. For a fair comparison, we train all models only using the synthetic dataset mentioned by Section \ref{sec:data} and test on the same noisy images. Compared with both reconstructive-based denoising methods and generative-based denoising methods, our model achieves considerable perceptual performance while maintaining comparable distortion and only needs 18 inference steps. Notably, pure diffusion-based denoising methods are lower than the reconstructed method (NAFNet) in terms of LPIPS perceptual metrics, which illustrates that pure diffusion-based denoising methods are prone to some artifacts. Although LDM achieves a lower NIQE metric, non-reference metrics are not always appropriate.

We further show the qualitative performance of our method in Figure \ref{fig:comparison-synthetic}. The proposed {\modelabbrname} inherits the advantages of both the fidelity from the reconstructive methods and the detail-preserving ability from the generative methods. By observing numeric and visual results, it can be seen that LPIPS is the perceptual metric that most closely aligns with human perception. For example, in the third scene, while LDM achieves the highest NIQE score, its generated texture is chaotic and far from visually satisfactory. Comparing our results with the NAFNet, it can be found that while the LPIPS is improved, the PNSR will slightly decrease, which corresponds to the perceptual-distortion trade-off.

\begin{table}[t]
\small
\vspace{-8pt}
\caption{Image denoising results on the SIDD dataset. Our {\modelabbrname} achieves the best LPIPS perceptual metric while maintaining comparable PSNR.}
\label{tab:sidd-result}
\centering
\resizebox{\columnwidth}{!}{
\begin{tabular}{@{}ccccccc@{}}
\toprule
\multicolumn{7}{c}{SIDD}                                                                                                                                                                                           \\ \midrule
Type                            & \multicolumn{1}{c|}{Model}                               & LPIPS↓            & NIQE↓              & PSNR↑              & \multicolumn{1}{c|}{SSIM↑}             & Inference Step \\ \midrule
N/A                             & \multicolumn{1}{c|}{Ground Truth}                        & 0.0               & 10.3890            & $+\infty$          & \multicolumn{1}{c|}{1.00}              & -              \\ \midrule
\multirow{4}{*}{Reconstructive} & \multicolumn{1}{c|}{CycleISP}                            & 0.2101            & 16.5107            & 39.2715            & \multicolumn{1}{c|}{0.9431}            & -              \\
                                & \multicolumn{1}{c|}{MAXIM}                               & 0.1895            & 16.6797            & 39.6786            & \multicolumn{1}{c|}{0.9469}            & -              \\
                                & \multicolumn{1}{c|}{Restormer}                           & 0.1978            & 16.8028            & 39.9282            & \multicolumn{1}{c|}{0.9469}            & -              \\
                                & \multicolumn{1}{c|}{NAFNet}                              & 0.1877            & 16.0501            & $\textbf{40.2110}$ & \multicolumn{1}{c|}{$\textbf{0.9496}$} & -              \\ \midrule
\multirow{2}{*}{Generative}     & \multicolumn{1}{c|}{DDRM}                                & 0.2879            & 18.1422            & 34.7716            & \multicolumn{1}{c|}{0.9148}            & 200            \\
                                & \multicolumn{1}{c|}{LDM}                                 & 0.0852            & 13.1519            & 34.7656            & \multicolumn{1}{c|}{0.9004}            & 200            \\ \midrule
\multirow{3}{*}{Both}           & \multicolumn{1}{c|}{NAFNet (with percepual loss)}        & 0.1790            & 16.2380            & 39.6472            & \multicolumn{1}{c|}{0.9457}            & -              \\
                                & \multicolumn{1}{c|}{PDASR(with GAN loss, P-D trade-off)} & 0.1298            & 12.1099            & 34.9306            & \multicolumn{1}{c|}{0.9194}            & -              \\
                                & \multicolumn{1}{c|}{RnG (Ours)}                          & $\textbf{0.0719}$ & $\textbf{11.8473}$ & 38.0291            & \multicolumn{1}{c|}{0.9256}            & 76             \\ \bottomrule
\end{tabular}
}
\end{table}

\begin{table}[t]
\small
\caption{Image denoising results on the DIV2K dataset. Our {\modelabbrname} achieves the best LPIPS perceptual metric while maintaining comparable PSNR.}
\label{tab:div2k-result}
\centering
\begin{tabular}{@{}ccccccc@{}}
\toprule
\multicolumn{7}{c}{DIV2K}                                                                                                                                                              \\ \midrule
Type                        & \multicolumn{1}{c|}{Model}        & LPIPS↓            & NIQE↓             & PSNR↑              & \multicolumn{1}{c|}{SSIM↑}             & Inference Step \\ \midrule
N/A                         & \multicolumn{1}{c|}{Ground Truth} & 0.0               & 7.9515            & $+\infty$          & \multicolumn{1}{c|}{1.00}              & -              \\ \midrule
Reconstructive              & \multicolumn{1}{c|}{NAFNet}       & 0.1308            & 10.6393           & $\textbf{33.1860}$ & \multicolumn{1}{c|}{$\textbf{0.8883}$} & -              \\ \midrule
\multirow{2}{*}{Generative} & \multicolumn{1}{c|}{DDRM}         & 0.1826            & 12.4311           & 30.4951            & \multicolumn{1}{c|}{0.8484}            & 200            \\
                            & \multicolumn{1}{c|}{LDM}          & 0.1656            & $\textbf{8.6438}$ & 26.8525            & \multicolumn{1}{c|}{0.7251}            & 200            \\
Both                        & RnG (Ours)                        & $\textbf{0.1040}$ & 9.1549            & 31.6678            & 0.8603                                 & 18             \\ \bottomrule
\end{tabular}
\vspace{-8pt}
\end{table}

\vspace{-8pt}
\subsection{Ablation Study and Analysis}
\vspace{-4pt}
\label{sec:ablation_framework}
\noindent\textbf{Framework.} To further illustrate the effectiveness of our framework, we conduct ablation experiments from three aspects: the training schemes, the initial denoising network, and the conditional input of the diffusion model. In terms of training strategies, three main strategies are mainly compared: joint training, intermediate supervised training, and two-stage training. In terms of the initial denoising network, we compared U-Net with NAFNet~\cite{chen2022simple} we currently used. In addition, we also explored the conditional input of the diffusion model using noisy images compared to the initial estimation result.

In Table~\ref{framework ablation}, we start from a simple U-Net architecture baseline, similar to~\cite{whang2022deblurring}, and gradually enable each of the aforementioned settings. All models are trained for 1M steps to ensure that the differences are not due to insufficient training. During the inference, we use 500 steps with a linear noise schedule, similar to~\cite{whang2022deblurring}. 
Through a detailed ablation study, we find that this setting reports the best performance: using a two-stage training scheme, the NAFNet as the reconstructive module, and the original noisy image as the conditional input of the diffusion model. This result proves the rationality of the framework design of the proposed RnG.

\begin{figure}[t]
  \centering
  \includegraphics[width=0.86\textwidth]{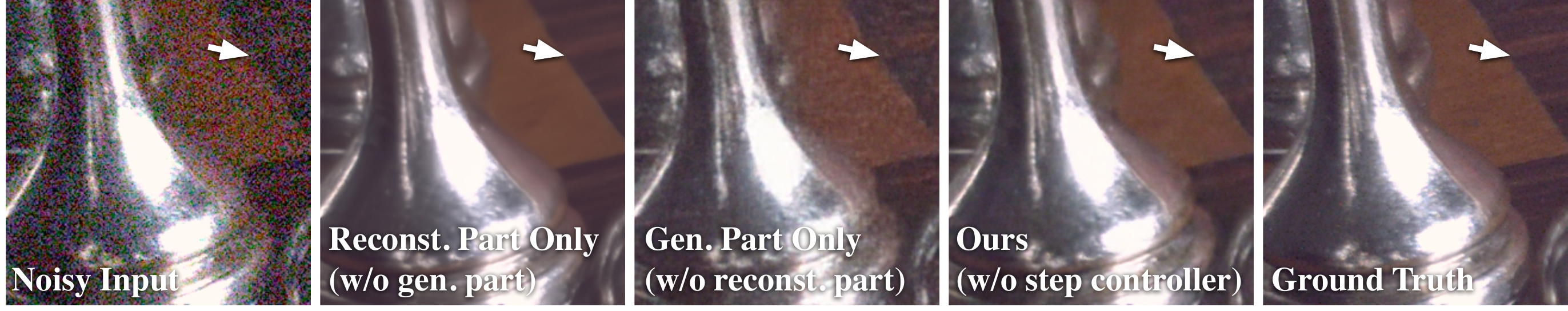}
  \caption{Visualization results of different components in reconstruct-and-generate diffusion model. We see that the reconstructive model predicts smooth results, the generative model predicts artifacts, and our {\modelabbrname} achieves better visual quality.}
  \label{fig:ablation-inner-comparison}
\vspace{-15pt}
\end{figure}

In addition, we compare our {\modelabbrname} with the reconstructive part only and the generative part only to show the effectiveness and plausibility of our designed framework. During inference, we use 2,000 steps with a linear noise schedule, similar to~\cite{saharia2022image}. As shown in Table~\ref{components ablation}, the reconstructive part only achieves the best distortion metric with lower perception, the generative part only can achieve good perception but cannot guarantee fidelity (lower PSNR), while our method can balance fidelity and perceptual quality, and achieve the best perceptual performance. More visual results can be found in Figure~\ref{fig:ablation-inner-comparison}. Our {\modelabbrname} can combine the advantages of reconstruction and generation to achieve a balance of perception and distortion.

\begin{figure}[h]
\centering
\includegraphics[width=0.85\linewidth,, cframe=gray!10!black 0.1mm]{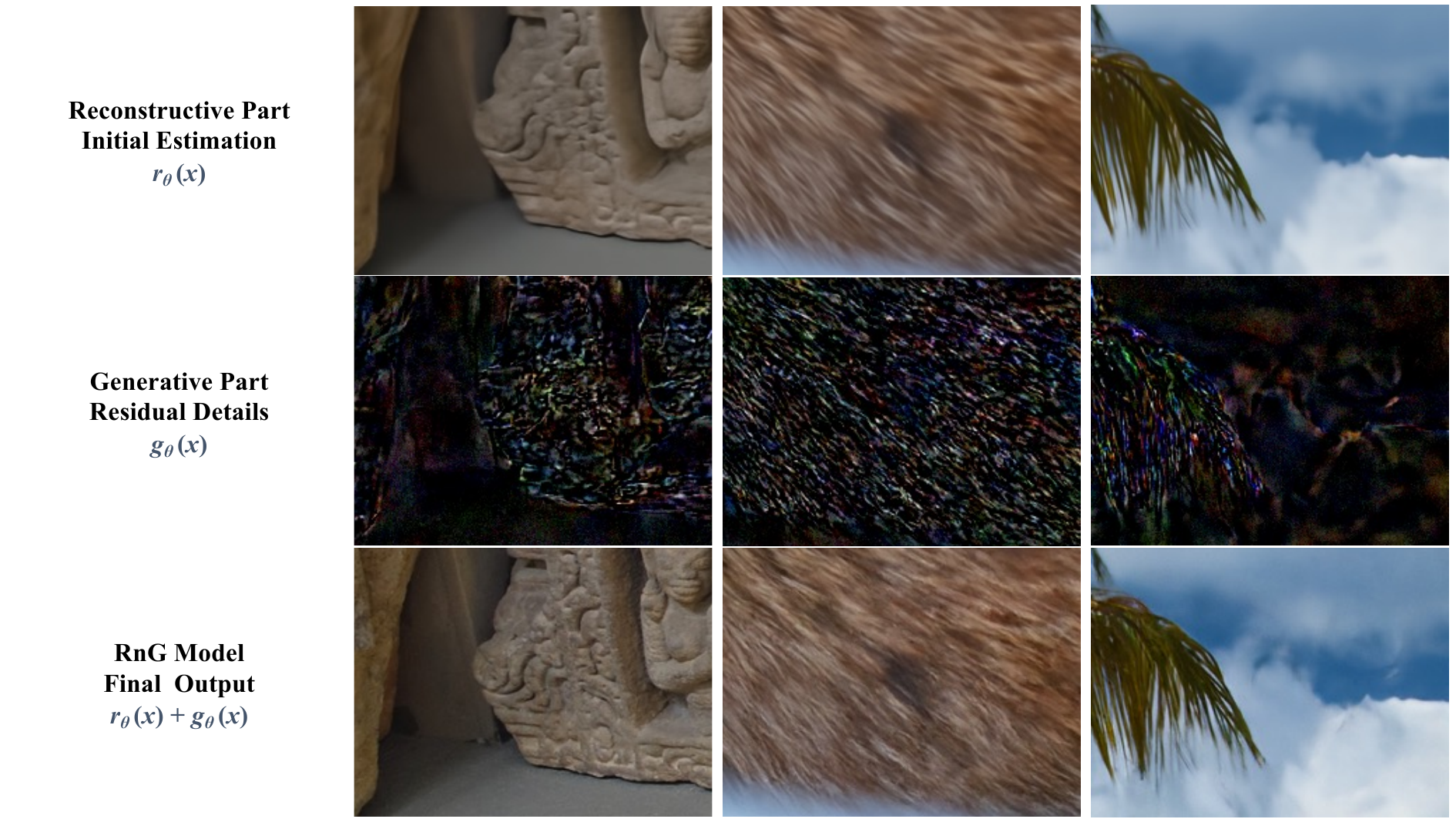}
\caption{The visualization of the high-frequent residual details produced by the generative part. The initial estimations produced by the reconstructive part as well as the final outputs of the RnG model are also presented. The generative part effectively brings more high-frequent details to the over-smoothed reconstructive output, achieving higher perceptual quality.}
\label{fig:delta_residual_detail}
\end{figure}

\noindent\textbf{What the Generative Part Learns.}
 We visualize the output of the generative part of the proposed RnG model on the DIV2K dataset to analyze what the generative part learns. Figure~\ref{fig:delta_residual_detail} shows that the residual detail information generated at each step within the diffusion model encompasses mainly high-frequency texture.

\noindent\textbf{Step Controller.} 
At last, we conduct an ablation study to demonstrate the effectiveness of the step controller module.
Table~\ref{big-model-step-ablation-on-synthetic} shows that our step controller module can bring significant improvements in both distortion and perception quality under the same inference step. These improvements demonstrate the importance of the step controller module. As shown in Figure~\ref{fig:demo-why-do-we-need-step-controller}, different texture regions have different requirements for steps, flat regions require fewer steps and textured regions require more steps.

\vspace{-4pt}
\begin{minipage}[t]{0.48\textwidth}
\centering
\captionsetup{type=table}
\small
\captionof{table}{Experimental results of ablation on components in {\modelabbrname} using SIDD dataset. Our {\modelabbrname} achieves the best perceptual metric while maintaining comparable PSNR.}
\label{components ablation}
\begin{tabular}{@{}ccc@{}}
\toprule
Model             & LPIPS↓            & PSNR↑              \\ \midrule
Reconstructive Part Only  & 0.1877            & $\textbf{40.2110}$ \\
Generative Part Only      & 0.0934            & 35.8663            \\
Ours (w/o step controller) & $\textbf{0.0891}$ & 37.8497            \\ \bottomrule
\end{tabular}
\end{minipage}\hspace{0.04\textwidth}
\begin{minipage}[t]{0.48\textwidth}
\centering
\captionsetup{type=table}
\small
\captionof{table}{Experimental results of ablation on step controller module using DIV2K dataset. Our method with a step controller module significantly outperforms our method with a fixed step under both perceptual and distortion metrics.}
\label{big-model-step-ablation-on-synthetic}
\begin{tabular}{@{}ccc@{}}
\toprule
Step Controller & LPIPS↓            & PSNR↑              \\ \midrule
                & 0.1078            & 31.0144            \\
$\checkmark$    & $\textbf{0.1040}$ & $\textbf{31.6678}$ \\ \bottomrule
\end{tabular}
\end{minipage}


\vspace{-12pt}
\section{Conclusion}
\vspace{-8pt}
In this paper, we combined the strengths of reconstruction and generation methods and presented the RnG diffusion model for image denoising, which focuses on pursuing a better balance between perception and distortion.
We further introduced a step controller to improve the perceptual quality of RnG and eliminate unnecessary computation costs. 
Our experimental results show that our method significantly improves perceptual quality compared to the current state-of-the-art methods while maintaining reasonable distortion quality. 
In future work, we would focus on reducing the computational complexity of the proposed methods since their computational cost is still expensive for edge devices, which is also a common problem of diffusion-based approaches.


\bibliography{neurips_2023}{}
\bibliographystyle{plain}  








\end{document}